\documentclass[11pt]{article}

\usepackage[final]{acl}

\usepackage{times}
\usepackage{latexsym}

\usepackage[T1]{fontenc}

\usepackage[utf8]{inputenc}

\usepackage{microtype}

\usepackage{inconsolata}

\usepackage{graphicx}
\usepackage[utf8]{inputenc}
\usepackage{booktabs} 
\usepackage[table]{xcolor} 
\usepackage{enumitem}
\usepackage{float}
\usepackage{amsmath,amssymb} 
\usepackage{dsfont} 
\usepackage{xspace}

\newcommand{\ours}[1][]{SWE-AGILE\ifx\relax#1\relax\else$_{\text{#1}}$\fi\xspace}

\newcommand{\hideit}[1]{}

\newcommand{\hui}[1]{\textcolor{blue}{\textbf{[Hui: #1]}}}

\author{
Shuquan Lian$^1$ \quad
Juncheng Liu$^2$ \quad
Yazhe Chen$^1$ \quad
Yuhong Chen$^1$ \quad
Hui Li$^1$
\\
$^1$Key Laboratory of Multimedia Trusted Perception and Efficient Computing Ministry of \\
Education of China, Xiamen University \\
$^2$ Microsoft \\
{\tt\small shuquanlian@stu.xmu.edu.cn, hui@xmu.edu.cn}
}

\title{SWE-AGILE: A Software Agent Framework for Efficiently Managing Dynamic Reasoning Context}

\begin{document}
\maketitle
\begin{abstract}
Prior representative ReAct-style approaches in autonomous Software Engineering (SWE) typically lack the explicit System-2 reasoning required for deep analysis and handling complex edge cases. While recent reasoning models demonstrate the potential of extended Chain-of-Thought (CoT), applying them to the multi-turn SWE task creates a fundamental dilemma: retaining full reasoning history leads to context explosion and ``Lost-in-the-Middle'' degradation, while discarding it would force the agent to redundantly re-reason at every step. To address these challenges, we propose SWE-AGILE, a novel software agent framework designed to bridge the gap between reasoning depth, efficiency, and context constraints. SWE-AGILE introduces a Dynamic Reasoning Context strategy, maintaining a ``sliding window'' of detailed reasoning for immediate continuity to prevent redundant re-analyzing, while compressing historical reasoning content into concise Reasoning Digests. Empirically, SWE-AGILE sets a new standard for 7B-8B models on SWE-Bench-Verified using only 2.2k trajectories and 896 tasks. Code is available at \url{https://github.com/KDEGroup/SWE-AGILE}.
\end{abstract}


\section{Introduction}
\label{sec:intro}

The blossoming of Large Language Models (LLMs) and Code Models~\citep{abs-2409-12186, 11273185} has revolutionized software engineering (SWE), enhancing efficiency in various tasks.
Particularly, there is a surge of works on SWE agents for autonomously navigating repositories, localizing bugs, and fixing bugs~\citep{DBLP:conf/iclr/JimenezYWYPPN24, DBLP:conf/nips/YangJWLYNP24}, i.e., the SWE task. 
The SWE task is difficult because it involves using tools, writing code, and debugging over multiple turns. 
These activities are closely connected, requiring heterogeneous reasoning capabilities.

\hideit{It is a mix of Deliberative Steps and Reflexive Steps, requiring heterogeneous reasoning demands. Deliberative steps—such as root cause analysis, hypothesis formulation, and interpreting conflicting logs—require deep, explicit System-2 reasoning to handle edge cases and maintain logical consistency. Conversely, Reflexive steps—such as syntax verification, directory navigation, or simple formatting—are routine operations that require minimal reasoning.}

Prior representative REACT-style~\citep{DBLP:conf/iclr/YaoZYDSN023} approaches like SWE-Dev~\citep{DBLP:conf/acl/WangHWTD25} and SWE-smith~\citep{yang2025swesmith} train models with limited context length (e.g., Qwen2.5~\citep{qwen2025qwen25technicalreport} and Qwen3~\citep{qwen3}) to generate actions alongside shallow thought traces. 
However, without explicit \emph{System-2 reasoning}~\cite{abs-2502-17419} that is more analytical and
deliberate, it is hard for these methods to perform deep analysis and handle edge cases correctly. 

Recent advancements in System-2 reasoning models, such as OpenAI o1~\citep{DBLP:journals/corr/abs-2412-16720} and DeepSeek-R1~\citep{DBLP:journals/corr/abs-2501-12948}, suggest that extending the Chain-of-Thought (CoT)~\citep{DBLP:conf/nips/Wei0SBIXCLZ22} length significantly enhances LLM's problem-solving capabilities. 
Attempting to harness this potential in automatic SWE, agentic scaffolds relying on powerful reasoning LLMs allow for long CoT during generation but discard these reasoning traces in the historical context, retaining only a concise description and the final action.
\hideit{Along this direction, various representative works are proposed, including but not limited to SWE-agent~\citep{DBLP:conf/nips/YangJWLYNP24}, OpenHands~\citep{DBLP:conf/iclr/0001LSXTZPSLSTL25} and SWE-Search~\citep{DBLP:conf/iclr/AntoniadesOZXGW25}
}

One key observation emerges from the recent progress in System-2 reasoning models is that discarding reasoning content upon the arrival of the second round of messages results in significant \emph{token inefficiency}, forcing the model to \emph{redundantly re-reason} through the entire problem for each subsequent tool call, as observed on DeepSeek-V3.2~\citep{deepseekai2025deepseekv32pushingfrontieropen}.
MiniMax M2~\citep{minmaxm2} also claims that Agents require \emph{Interleaved Thinking} and retaining the full session history, including the reasoning content. 
This approach is made possible by its massive context capacity of 204,800 tokens.

\begin{figure}[t]
    \centering
    \includegraphics[width=0.95\linewidth]{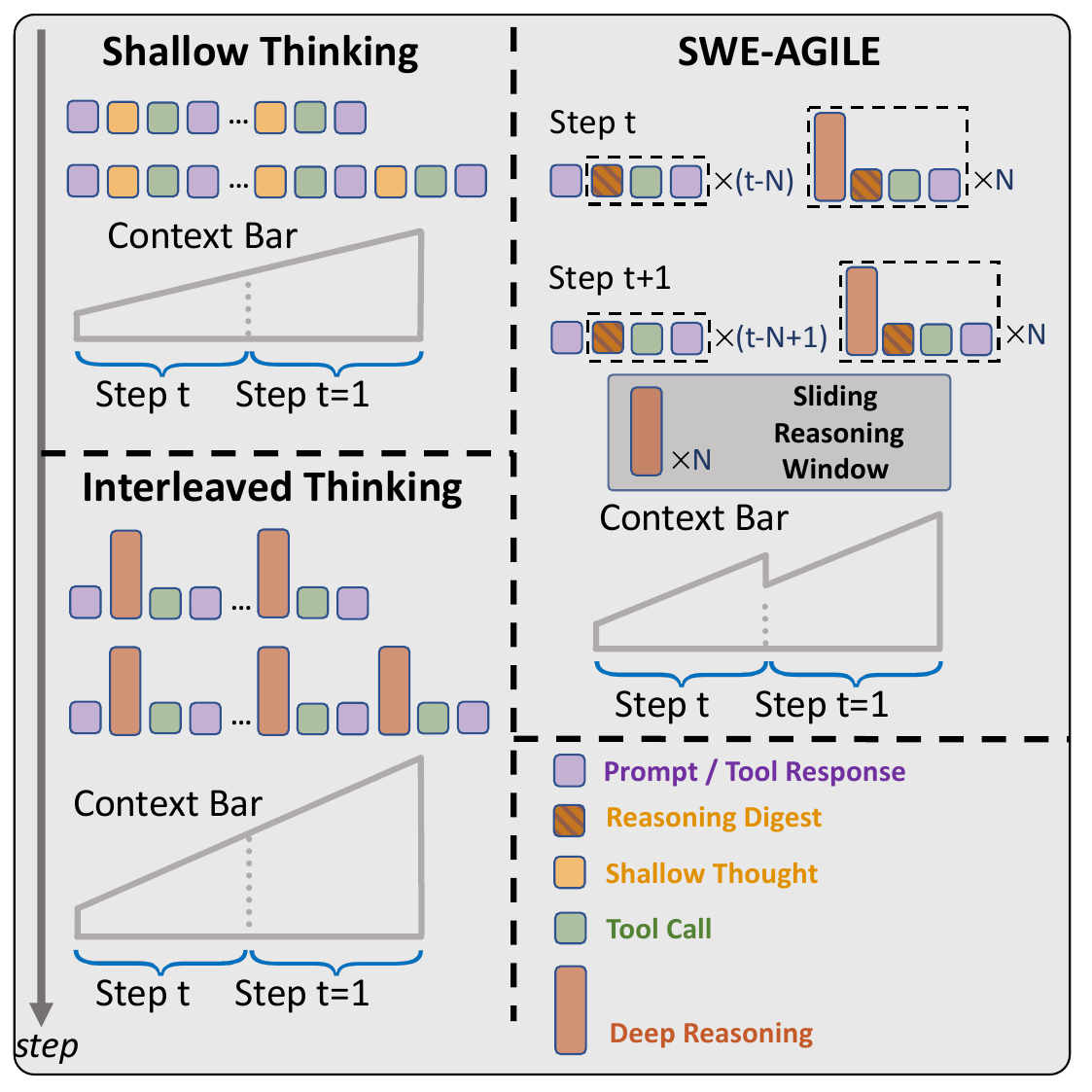}
    \vspace{-10pt}
    \caption{A comparison of context growth patterns in the multi-turn SWE task: (1) Shallow Thinking maintains low context cost but lacks reasoning depth. (2) Interleaved Thinking enables deep System-2 reasoning but suffers from rapid, linear context growth (steep context bar). (3) \ours enables a sustainable ``Sawtooth'' growth pattern. Within the \textit{Sliding Reasoning Window}, the agent engages in deep reasoning (steep slope similar to Interleaved Thinking); however, as steps progress, historical thoughts are compressed into concise \textit{Reasoning Digests}.}
    \vspace{-10pt}
    \label{fig:context}
\end{figure}

However, indiscriminately applying full-history Interleaved Thinking to the multi-turn SWE task presents fundamental scalability and efficiency challenges. 
SWE agents typically engage in frequent interactions involving extensive code retrievals and verbose execution logs. 
When retaining the long CoT from every step, \emph{the context window expands rapidly}.
This creates two critical issues. 
First, even for models capable of processing long inputs, performance often degrades as context length increases, a phenomenon known as ``Lost-in-the-Middle'' or attention dilution~\citep{DBLP:journals/tacl/LiuLHPBPL24}, where the model struggles to retrieve relevant information from the middle of a long sequence. 
Second, processing such massive sequences demands \emph{excessive GPU memory} and drastically reduces \emph{training speed} due to computational overhead. 


To address the above issues, we propose a novel framework \ours designed to bridge the gap between reasoning depth, efficiency, and context constraints in in the multi-turn SWE tasks. Fig.~\ref{fig:context} contrasts our approach with Shallow Thinking and Interleaved Thinking at the level of inference.
The main contributions of this work are summarized as follows:
\begin{itemize}[leftmargin=10pt,topsep=1pt,itemsep=0.2pt] 

\item \textbf{Dynamic Reasoning Context:} We propose a hybrid dynamic context management strategy that compresses every historical reasoning content into a concise reasoning digest for long-term retention, while preserving a ``Last-N-Steps'' sliding reasoning window of long CoT to maintain cognitive continuity in the working context and avoid redundantly re-analyzing the global state at every turn.

\item \textbf{Trajectory Snapshot Training:}
We introduce a snapshot-based training objective to address the misalignment between standard training and dynamic inference. By decomposing trajectories into discrete snapshots with context-aware masking, we force the model to learn under the constraints of dynamic context at runtime.

\item \textbf{Backfilling Data Synthesis:} We develop a ``Hindsight Backfill'' pipeline that augments successful trajectories with detailed reasoning content and reasoning digests. This process leverages future ground-truth actions to synthesize high-quality, format-compliant training data tailored to our dynamic context constraints.

\item \textbf{Compression-Aware Optimization:} We design a trajectory-level Compression Rate Reward within the Reinforcement Learning with Verifiable Rewards (RLVR) process. This mechanism incentivizes the model to generate sufficiently detailed reasoning for problem-solving while maximizing the conciseness of the digests, effectively balancing performance with token efficiency.

\end{itemize}


Empirically, we validate \ours on the SWE-Bench Verified benchmark. Utilizing merely \textbf{2.2k} training trajectories, \ours achieves a \textbf{24.1\%} success rate, surpassing all existing 7B/8B baselines. We attribute this success to the effectiveness of our proposed paradigm: by fundamentally resolving the conflict between reasoning depth and context constraints, \ours effectively elicits latent System-2 reasoning ability.

\section{Our Method}

\begin{figure*}[t]
\centering
\includegraphics[width=0.97\textwidth]{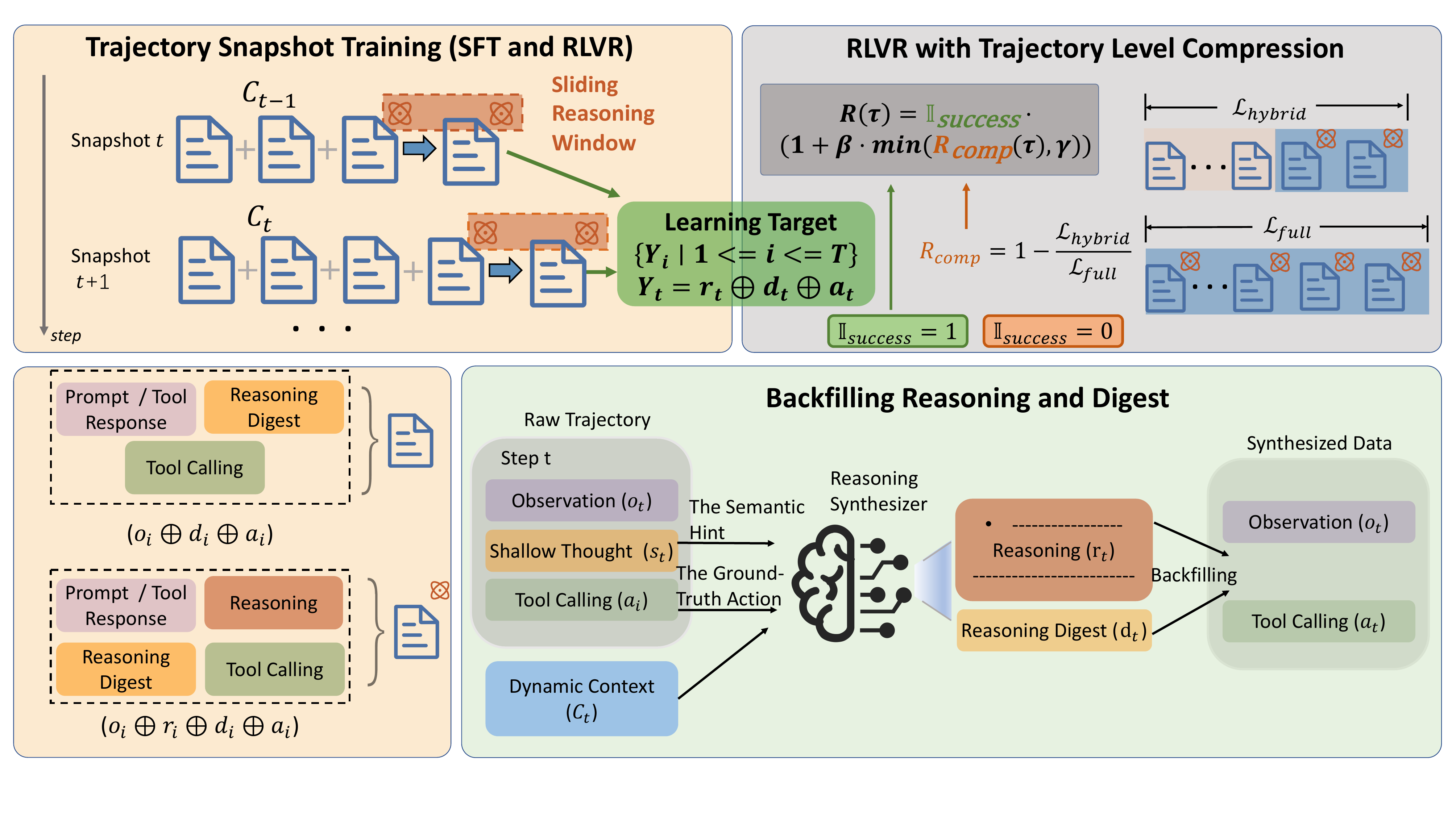} 
\vspace{-20pt}
\caption{
    Overview of \ours. 
    (1) \emph{Trajectory Snapshot Training}: We decompose long trajectories into discrete snapshots. In each snapshot, historical reasoning traces outside the \textit{Sliding Reasoning Window} are replaced by digests ($d_i$) and masked from the loss, forcing the model to learn the target $Y_t$ (Reasoning $r_t \to$ Digest $d_t \to$ Action $a_t$) based on a compact context $C_t$.
    (2) \emph{Backfilling Reasoning and Digest}: We synthesize high-quality training data by employing a reasoning model to augment raw trajectories. Conditioned on the ground-truth future action, the semantic hint and the dynamic context, the model backfills detailed reasoning ($r_t$) and concise digest ($d_t$).
    (3) \emph{RLVR with Trajectory Level Compression:} We introduce a compression-aware reward function. The model is incentivized to maximize the reduction ratio between the actual hybrid context ($\mathcal{L}_{hybrid}$) and the hypothetical full context ($\mathcal{L}_{full}$), thereby learning to efficiently compress historical information without sacrificing task success ($\mathbb{I}_{success}$). 
}
\vspace{-10pt}
\label{fig:training}
\end{figure*}

Fig.~\ref{fig:training} provides an overview of \ours. To enable deep reasoning within a sustainable context window, our framework is realized through three parts: (1) Trajectory Snapshot Training (Sec.~\ref{sec:snapshot}); (2) Backfilling Reasoning and Digest (Sec.~\ref{sec:backfill}); and (3) RLVR with Trajectory Level Compression (Sec.~\ref{sec:rlvr}).

\hideit{
LightThinker~\citep{zhang-etal-2025-lightthinker} trains the model to predict a trigger token when it encounters a thought boundary (such as ``\n\n'') or after generating a fixed number of reasoning tokens. When the trigger token appears, the model performs a compression step: it compresses the hidden states of the preceding thought via the attention-based aggregation mechanism of the gist tokens, producing a fixed number (K) of gist-token hidden states. These K gist tokens are then written into the KV cache, and the original verbose tokens are discarded.

InftyThink~\citep{DBLP:journals/corr/abs-2503-06692} proposes an iterative long-reasoning framework in which a model solves a problem through a sequence of reasoning segments. Each segment consists of accumulated sentences constrained by a fixed token budget. After generating a segment, the model compresses it together with the previous summary into a new text-based summary, which then replaces all earlier content. The model then produces the next reasoning segment, and this new segment is again compressed with the updated summary. This process repeats iteratively.

While the two mentioned methods focus on dynamic compression within the internal reasoning process of a long single-turn CoT, we propose a novel paradigm tailored for scenarios requiring multi-turn environmental interactions, such as SWE-Agent. Notably, our method is orthogonal to the previously discussed CoT compression techniques and can be deployed in conjunction with them.
}

\subsection{Dynamic Reasoning Context and Trajectory Snapshot Training}
\label{sec:snapshot}



\ours formalizes the agent's interaction as a sequence of steps $t = \{1, 2, 3, \cdots, T\}$. 
At each step, the generation and the context management operate as follows:
for step $t$, the model is required to generate a composite response $y_t$, consisting of three strictly ordered components:
\begin{equation}
\label{eq:target_def}
    Y_t = r_t \oplus d_t \oplus a_t
\end{equation}
where $r_t$ is the detailed reasoning for analyzing the current state, $d_t$ is the reasoning digest (i.e., a concise digest of $r_t$ generated immediately thereafter), and $a_t$ is the executable action.

To generate $Y_t$, the model conditions on a hybrid context $C_t$. 
Let $N$ denote the size of the sliding reasoning window. 
The context $C_t$ is formulated as the concatenation of long-term history with condensed reasoning and a detailed reasoning sliding window. Specifically, while the full history of environmental interactions (observations and actions) is always retained, the reasoning traces ($r$) outside the sliding window are replaced by their concise digests ($d$):
\begin{equation}
\label{eq:context_def}
\begin{aligned}
C_t = &{\Big[ \bigcup_{i=0}^{t-N-1} (o_i, d_i, a_i) \Big]} \oplus\\
&{\Big[ \bigcup_{j=t-N}^{t-1} (o_j, r_j, d_j, a_j) \Big]} \oplus o_t
\end{aligned}
\end{equation}
where $o_t$ is the observation of the tool calling result.

Our approach differs from standard interaction summarization (e.g., LangChain's ConversationSummaryBufferMemory and MemGPT's working context)~\citep{packer2024memgptllmsoperatingsystems}. LangChain performs incremental summarization of the previous dialogue history, resulting in a single, growing summary paragraph. MemGPT utilizes a fixed-size block for unstructured text writable via function calls. In contrast:
\begin{itemize}[leftmargin=10pt,topsep=1pt,itemsep=0.2pt] 

\item \textbf{Targeted Reasoning Compression:} : Our Reasoning Digests specifically target the reasoning traces rather than the general interaction history.

\item \textbf{Structured Modularity:} Unlike a monolithic summary, our per-step digests remain distinct entities. This structured format not only \emph{better aligns with pre-trained LLM behaviors}, but also mitigates the compounding risk of error propagation—a prevalent challenge in sequential and hierarchical LLM reasoning~\citep{ma2026hcrellmbasedhierarchicalclassification}.
\end{itemize}
 
Standard SFT and RL typically treat a multi-turn trajectory as a single contiguous sequence. 
However, \ours's inference imposes a strict dynamic visibility constraint: \emph{while environmental history (observations and actions) remains fully visible, detailed reasoning traces ($r_t$) are transient}. 
They are retained only within a sliding window ($N$ steps) for cognitive continuity before being permanently replaced by digests ($d_t$).
Training under the standard contiguous sequence assumption violates this dynamic context setting, as it allows the model to attend to all historical reasoning traces ($r_1, r_2, \dots$), which are explicitly hidden during inference.
Therefore, applying such a dynamic context strategy brings the misalignment between training and inference. 
Implementing complex, dynamic attention masks to retroactively hide previous reasoning tokens is engineering-intensive, disrupting standard efficiency optimizations like FlashAttention~\citep{DBLP:conf/nips/DaoFERR22}, and introducing significant training overhead.

To align the training process with the inference process, we decompose each full trajectory $\tau$ of length $T$ into a set of discrete Trajectory Snapshots $\mathcal{S} = \{ (C_t, Y_t) \}_{t=1}^{T}$.
Each snapshot represents the agent's specific ``world view'' at step $t$, optimized via a focused masking strategy:
\begin{itemize}[leftmargin=10pt,topsep=1pt,itemsep=0.2pt]

\item \textbf{Frozen Snapshot Context (Mask=0):} The input $C_t$ (Eq.~\ref{eq:context_def}) simulates the runtime state where historical reasoning outside the sliding window is already compressed. These tokens are visible to attention but excluded from the loss, acting as a fixed prompt.

\item \textbf{Active Target (Mask=1):} The target $Y_t$ (Eq.~\ref{eq:target_def}) comprises the current reasoning, reasoning digest, and action. This is the sole learnable segment for the snapshot.

\end{itemize}
This decomposition ensures a rolling optimization: every reasoning trace $r_t$ is optimized exactly once in the active target $Y_t$ and subsequently serves as a compressed reasoning digest in future snapshots, effectively resolving the training-inference mismatch.

The necessity of the trajectory snapshots training can be also found in recent studies of context management. AgentFold~\citep{DBLP:journals/corr/abs-2510-24699} executes a ``folding'' operation, which manages its historical trajectory at multiple scales: it can perform granular condensations to preserve vital, fine-grained details, or deep consolidations to abstract away entire multi-step subtasks. 
Context-Folding~\citep{DBLP:journals/corr/abs-2510-11967} procedurally branches into a sub-trajectory to handle a subtask and then folds it upon completion, collapsing the intermediate steps while retaining a concise summary of the outcome. 
They both treat a trajectory as multiple training examples due to context modification. 

\subsection{Backfilling Reasoning and Digest}
\label{sec:backfill}

We further introduce a hindsight backfill pipeline to synthesize reasoning and digests based on existing successful trajectories to support the training of \ours, inspired by ActRe~\citep{yang2024react} and UI-TARS~\citep{DBLP:journals/corr/abs-2501-12326} on GUI agent that address the lack of explicit reasoning in GUI action traces through annotating intermediate ``thoughts''. 
However, unlike recorded GUI traces which are often purely action-driven, existing SWE trajectories typically contain shallow natural language responses. 
Our pipeline upgrades these sparse signals into explicit System-2 reasoning traces ($r_t$) and reasoning digests ($d_t$), bridging the gap between shallow heuristics and deep problem-solving.

The pipeline operates on a raw trajectory $\tau = \{(o_1, s_1, a_1), \dots, (o_T, s_T, a_T)\}$, where $s_t$ is the shallow thought. 
We employ a strong reasoning model as the reasoning synthesizer to backfill the detailed reasoning traces of the trajectory step-by-step.

For each step $t$, the synthesizer generates the reasoning trace $r_t$ and reasoning digest $d_t$ conditioned on three critical inputs:
\begin{itemize}[leftmargin=10pt,topsep=1pt,itemsep=0.2pt] 

    \item \textbf{The Ground-Truth Action $a_t$:} Unlike standard inference, the model is provided with the future action, allowing it to backfill a CoT that inevitably leads to the correct decision.
    
    \item \textbf{The Semantic Hint $s_t$:} The original shallow thought is provided to preserve the agent's initial intent, ensuring the synthesized reasoning remains grounded in the trajectory's logic.
    
    \item \textbf{The Dynamic Context $C_t$:} We strictly simulate the inference-time visibility constraints defined in Eq.~(\ref{eq:context_def}). It includes the full history of environmental interactions (observations and actions) but applies dynamic compression to reasoning: thoughts prior to the sliding window are replaced by digests $d$, while detailed thoughts $r$ are retained only for the most recent $N$ steps. Crucially, by exposing the immediate history of detailed thoughts, we enable cognitive continuity: the model builds incrementally upon its recent reasoning process rather than redundantly re-analyzing the global state at every turn.

\end{itemize}
Ultimately, each step is reformatted into a structured tuple $(o_t, r_t, d_t, a_t)$, where the shallow thought $s$ is replaced with detailed reasoning $r$ and reasoning digest $d$. 

We prioritize this backfilling strategy over direct RLVR or rejection sampling for three reasons: 
\begin{itemize}[leftmargin=10pt,topsep=1pt,itemsep=0.2pt] 

\item \textbf{Adaptive Reasoning Efficiency:} In the SWE task, the required depth of reasoning varies widely across steps. Some steps are routine operational actions (e.g., executing a script as previously planned or simple navigation) that require only surface-level intent verification. While other steps, such as analyzing a confusing error message, designing a new function structure or figuring out why a bug occurred, require sustained System-2 reasoning to handle ambiguity. We categorize them into Reflexive Steps and Deliberative Steps. Backfilling reasoning enables more controllable reasoning depth compared to raw trajectory collecting.

\item \textbf{Format Enforcement:} Pre-trained models' generation often fails to strictly adhere to the ``Reasoning $\rightarrow$ Digest $\rightarrow$ Action'' format during multi-turn interactions. 
Backfilling reasoning and digest allow explicitly enforcing this format, creating a stable starting checkpoint for later RLVR.

\item \textbf{Data Scalability:} 
Directly collecting new trajectories using our paradigm from scratch is computationally expensive due to the high cost of environment execution and the low success rate. 
To alleviate this issue, we leverage existing successful trajectories, allowing rapidly synthesizing the data by simply annotating gold trajectories and avoiding the need for extensive exploration.

\end{itemize}

\subsection{Optimization via RLVR with Reasoning Compression}
\label{sec:rlvr}

After SFT, we employ RLVR to optimize the policy. The objective is to increase the task success rate while decoupling reasoning depth from context cost: the agent should learn to expand its reasoning ($r_t$) sufficiently to solve complex problems (Deliberative Steps), while minimizing the permanent context via concise Reasoning Digests ($d_t$).

\vspace{5pt}
\noindent\textbf{Trajectory-Level Compression Rate.} We first introduce a metric to quantify the efficiency of context management. 
Let $|\tau_t|$ denote the token length of the complete interaction tuple $(o_t, r_t, d_t, a_t)$ at step $t$. $\mathcal{L}_{full} = \sum_{t} |\tau_t|$ is the hypothetical total context size if the full history (including all reasoning $r_t$) were retained. 
$\mathcal{L}_{hybrid}$ is the actual context size under our dynamic policy, where $r_t$ is pruned from the history outside the sliding window.
The trajectory-level compression rate is defined as the global reduction ratio:
\begin{equation}
R_{comp} = 1 - \frac{\mathcal{L}_{hybrid}}{\mathcal{L}_{full}}.
\end{equation}
This metric reflects the percentage of total context memory saved.

\vspace{5pt}
\noindent\textbf{Reward Function.} 
The overall reward $R(\tau)$ conditions efficiency on effectiveness:
\begin{equation}
R(\tau) = \mathbb{I}_{\text{success}} \cdot \Big( 1 + \beta \cdot \min(R_{\text{comp}}(\tau), \gamma) \Big),
\end{equation}
where $\mathbb{I}_{\text{success}} \in \{0, 1\}$ denotes task success, $\beta$ denotes the weight of the compression reward, and $\gamma$ is a clipping threshold.
The clipping mechanism prevents the model from artificially bloating reasoning traces ($r_t$) merely to inflate the denominator of $R_{\text{comp}}$ beyond the saturation point.

The multiplicative gating $\mathbb{I}_{\text{success}}$ ensures that compression rewards are added only on successful trajectories, effectively preventing the model from trading correctness for compression scores.

\vspace{5pt}
\noindent\textbf{Global vs. Local Compression Rate: Handling Heterogeneous Step Complexity.} 
A critical design choice is utilizing a global trajectory-level metric rather than an average of step-wise compression ratios (e.g., $\frac{1}{T}\sum (1 - \frac{{len}(d_t)}{{len}(r_t)})$).
This design choice is driven by the categorization of Reflexive Step and Deliberative Steps of the SWE task described in Sec.~\ref{sec:backfill}:

\begin{itemize}[leftmargin=10pt,topsep=1pt,itemsep=0.2pt]

\item \textbf{Robustness to Reflexive Steps:} In Reflexive Steps, the necessary reasoning $r_t$ is naturally brief, often resulting in a reasoning digest $d_t$ of comparable length (i.e., $|r_t| \approx |d_t|$). This yields a near-zero local compression score. A step-wise objective would incentivize the model to ``reward hack'' by inflating $r_t$ with redundant tokens during these simple steps merely to increase the denominator and improve the local ratio.

\item \textbf{Incentivizing Depth in Deliberative Steps:}
The global compression rate is tolerant of low compression in Reflexive Steps (which contribute minimally to the total sums). Instead, it drives the model to focus its optimization efforts on Deliberative Steps, where $|r_t|$ is large and the potential for substantial context saving ($|r_t| \rightarrow |d_t|$) is high.

\end{itemize}

Consequently, this optimization process establishes a dynamic balance between reasoning depth and context compression. 
By decoupling the transient reasoning overhead (generating $r_t$) from the permanent context retention (storing $d_t$), the model learns an adaptive strategy: 
it behaves as a deep thinker during complex problem-solving moments to ensure $\mathbb{I}_{\text{success}}$, while acting as a concise distiller of its historical thoughts to maximize $R_{\text{comp}}$, thereby achieving competitive performance and token efficiency.

\section{Experiment}
\label{sec:experiment}

\subsection{Experiment Setup}


\vspace{5pt}
\noindent\textbf{Dataset.} 
Our training pipeline consists of two stages utilizing distinct data sources. 
For the Cold-Start SFT phase, we use a high-quality subset of 2.2k trajectories from the SWE-Dev dataset~\citep{DBLP:conf/acl/WangHWTD25} (originally 19.3k). 
Since the original data lacks explicit System-2 reasoning, we apply our backfilling data synthesis pipeline (Sec.~\ref{sec:backfill}) to augment these trajectories using Qwen3-235B-A22B-Instruct-2507\footnote{\url{https://huggingface.co/Qwen/Qwen3-235B-A22B-Instruct-2507}}.
\hideit{with dense reasoning traces ($r_t$) and reasoning digests ($d_t$).}
We further collect 200 trajectories via rejection sampling using Qwen3-235B-A22B-Thinking-2507\footnote{\url{https://huggingface.co/Qwen/Qwen3-235B-A22B-Thinking-2507}}. 
This small batch utilizes the exact same paradigm and scaffold as the subsequent RLVR phase to minimize distribution shift.

We compare all methods on the SWE-Bench-Verified~\citep{DBLP:conf/iclr/JimenezYWYPPN24} benchmark, which evaluates AI systems on their ability to solve 500 software issues from 12 real world GitHub repositories.

\vspace{5pt}
\noindent\textbf{Base Model.} 
Unlike prior research (e.g., R2E-Gym~\citep{jain2025regym} that utilizes the coding-specialized Qwen2.5-Coder~\citep{qwen2.5-coder}), we select Qwen3~\citep{qwen3} as our base model. 
This choice is driven by our research objective to explore the potential of \ours. 
A base model with strong reasoning capabilities serves as a more suitable subject for this exploration than a model strictly optimized for coding capabilities without thinking abilities.

\vspace{5pt}
\noindent\textbf{Implementation Details.} 
Our agentic scaffold is built upon R2E-Gym. 
We make slight modifications on the prompt to enforce the proposed ``Reasoning $\to$ Digest $\to$ Action'' workflow. See more detail in Appendix~\ref{app:scaffold}.
For the RLVR phase, we utilize 896 diverse tasks from the R2E-Gym subset environment. 
At the SFT stage, detailed hyperparameters are provided in Appendix~\ref{app:hyper_parameters}. During the rollout phase of RLVR, we set the maximum number of steps to 50, and a strict maximum number of tokens every response generated to 4096 to avoid too verbal reasoning. 
Any rollout that triggers one of the conditions in max steps, max tokens per response, max context, trajectory timeout or submit will terminate. 
We use DAPO~\citep{DBLP:journals/corr/abs-2503-14476} algorithm to optimize policy, and detailed hyperparameters are provided in Appendix~\ref{app:hyper_parameters}.
We use XML-based tool calling format (see Appendix~\ref{app:toolcall} for discussion).

\vspace{5pt}
\noindent\textbf{Evaluation Settings.} 
We limit the maximum number of steps to 60, the maximum number of context tokens to 65536. We run each evaluation 2 times and report the mean value of metrics.

\subsection{Overall Performance}

Tab.~\ref{tab:main_results} summarizes the performance of \ours compared to state-of-the-art open-source and closed-source models on SWE-Bench-Verified. 
From the results, we have the following observations.

Our method \ours, utilizing the Qwen3-8B model, establishes a new performance standard for models in the 7B-8B parameter class.
Starting from a general-purpose Qwen3-8B base, \ours (SFT) achieves a success rate of \textbf{21.45\%}, representing a substantial \textbf{35.5\% relative improvement} over the base model (15.83\%). This verifies the efficacy of \ours paradigm in eliciting System-2 reasoning capabilities even in 8B models.
Remarkably, \ours (SFT) achieves this performance using only \textbf{2.2k training trajectories}, a mere \textbf{11\%} of the 19.3k dataset utilized by SWE-Dev. 
With the integration of our compression-aware RLVR, \ours further elevates the success rate to \textbf{24.05\%}, outperforming all reported baselines of comparable size.
Notably, despite being an 8B model, \ours surpasses the Qwen3-based SkyRL-Agent-v0-14B (21.6\%). Additionally, we explored the scalability of our method on larger models. While computational constraints limited our ability to perform full RLVR on a 14B model, applying our SFT pipeline to Qwen3-14B yielded a success rate of 30.06\% on SWE-Bench-Verified, significantly surpassing existing 14B baselines. We also evaluated SWE-AGILE-8B on SWE-Bench Lite, where it achieved a success rate of 14.77\%, outperforming comparable baselines such as SWE-smith-7B (11.7\%) and R2E-Gym (11.0\%).

\definecolor{highlightblue}{RGB}{240, 240, 255}

\begin{table*}[t]
    \scalebox{0.8}{
    \renewcommand{\arraystretch}{0.83} 
     \centering
    \begin{tabular}{llll|c}
        \toprule
        Approach & Base Model & Scaffold & Data & \textsc{Success Rate} \\
        \midrule
        \multicolumn{5}{c}{\textit{Closed Weight Models}} \\
        \midrule
        OpenHands~\citep{DBLP:conf/iclr/0001LSXTZPSLSTL25} & GPT-5 & OpenHands & - & 71.80 \\
        OpenHands & Claude 4 Sonnet & OpenHands & - & 70.40 \\
        SWE-agent~\citep{DBLP:conf/nips/YangJWLYNP24} & Claude 4 Sonnet & SWE-agent & - & 66.6 \\
        Agentless-1.5~\citep{DBLP:journals/pacmse/XiaDDZ25} & GPT-4o & Agentless & - & 38.8 \\
        SWE-RL~\citep{wei2025swerl}  & Llama-3.3-70B & Agentless Mini & - & 41.0 \\
        \midrule
        \multicolumn{5}{c}{\textit{Open Weight Models}} \\
        \midrule
        \multicolumn{1}{c}{\textit{Lager than 32B}} \\
        \midrule
        Qwen3-Coder & Qwen3-Coder-480B & OpenHands & - & 69.6 \\
        Kimi-K2 & Kimi-K2-1T & OpenHands & - & 65.4 \\
        GLM-4.5 & GLM-4.5-355B & OpenHands & - & 64.2 \\
        SWE-fixer~\citep{DBLP:conf/acl/XieLGDLZ025} & Qwen2.5-72B & Pipline & - & 32.8 \\
        Kimi-Dev~\citep{DBLP:journals/corr/abs-2509-23045} & Qwen 2.5-72B & SWE-Agent & 150B tokens + & 48.6 \\
        CodeFuse-CGM~\citep{tao2025code} & Qwen2.5-72B & Graph RAG & 200k issue-patch pairs & 50.4 \\
        \midrule
        \multicolumn{1}{c}{\textit{32B/30B}} \\
        \midrule
        SWE-Gym~\citep{pan2025training} & Qwen2.5-Coder-32B & OpenHands & - & 20.6 \\
        R2EGym~\citep{jain2025regym} & Qwen2.5-Coder-32B & R2EGym & 3.3k trajectories & 34.4 \\
        SWE-Dev~\citep{DBLP:conf/acl/WangHWTD25} & Qwen2.5-Coder-32B & OpenHands & 19.3k trajectories & 36.6 \\
        Skywork-SWE~\citep{DBLP:journals/corr/abs-2506-19290} & Qwen2.5-Coder & OpenHands & 8k trajectories & 38.0 \\
        SWE-smith~\citep{yang2025swesmith} & Qwen2.5-Coder-32B & SWE-agent & 5k trajectories & 40.2 \\
        DeepSWE~\citep{deepswe2025} & Qwen3-32B & R2EGym & 4.5k SWE tasks & 42.2 \\
        ENTROPO-KTO~\citep{DBLP:journals/corr/abs-2509-12434} & Qwen3-Coder-30B & R2EGym & - & 49.3 \\
        \midrule
        \multicolumn{1}{c}{\textit{14B}} \\
        \midrule
        SWE-Gym & Qwen2.5-Coder-14B & OpenHands & - & 16.4 \\
        SkyRL-Agent-v0~\citep{cao2025skyrl} & Qwen3-14B & OpenHands & - & 21.6 \\
        R2EGym & Qwen2.5-Coder-14B & R2EGym & 3.3k trajectories & 26.8 \\ 
        \ours (SFT) & Qwen3-14B & R2EGym & 2.2k trajectories & 30.06 \\
        \midrule
        \multicolumn{1}{c}{\textit{7B/8B}} \\
        \midrule
        SWE-Gym & Qwen2.5-Coder-7B & OpenHands & 491 trajectories & 10.6 \\
        SWE-smith & Qwen2.5-Coder-7B & SWE-agent & 5k trajectories & 15.2 \\
        R2EGym & Qwen2.5-Coder-7B & R2EGym & 3.3k trajectories & 19.0 \\
        SWE-Dev & Qwen2.5-Coder-7B & OpenHands & 19.3k trajectories & 23.4 \\
        Qwen3 & Qwen3-8B & R2EGym & - & 15.83 \\
        \ours (SFT) & Qwen3-8B & R2EGym & 2.2k trajectories & 21.45 \\
        \ours (SFT+RL) & Qwen3-8B & R2EGym & + 896 SWE tasks & 24.1 \\
        \bottomrule
    \end{tabular}
    }
    \vspace{-5pt}
    \caption{Performance Comparison on SWE-Bench-Verified benchmark.}
    \vspace{-5pt}
    \label{tab:main_results}
\end{table*}

\hideit{
\hui{?}

Findings 1: With RL processes, average step number per trajectory increases.
}
\hideit{
\subsection{Reasoning Window Analysis}

We investigate the impact of the Sliding Reasoning Window size.
}

\subsection{Analysis of Dynamic Reasoning Context}
\label{sec:analysis}

We utilize the Qwen3-8B model to conduct a controlled analysis of reasoning context management and training paradigms. The results, detailing accuracy, average steps, and token consumption, are presented in Tab.~\ref{tab:ablation_metrics}.

\begin{table}[t]
    \scalebox{0.65}{
        \centering
        \renewcommand{\arraystretch}{0.9}
        \begin{tabular}{l|c|c}
            \toprule
            \textbf{Method} & \textbf{Avg Steps} & \textbf{Success Rate (\%)} \\
            \midrule
            \multicolumn{3}{l}{\textit{Inference Baselines (Base Model)}} \\
            (1) Disable Thinking & 43.95 & 0.03 \\
            (2) Interleavd Thinking & 26.08 & 12.42 \\
            (3) Current-Step Thinking & 15.77 & 15.83 \\
            \midrule
            \multicolumn{3}{l}{\textit{SFT}} \\
            (4) Standard SFT (w/o Backfilling) & 23.85 & 14.83 \\
            (5) \textbf{\ours (SFT)} & 21.00 & 21.45 \\
            \midrule
            \multicolumn{3}{l}{\textit{RLVR}} \\
            (6) Standard SFT+RL & 18.89 & 16.03 \\
            (7) \textbf{\ours (SFT+RL w/o CR)} & 22.66 & 23.45 \\
            (8) \textbf{\ours (SFT+RL)} & 24.59 & 24.05 \\
            \bottomrule
        \end{tabular}
    }
    \vspace{-5pt}    
    \caption{
        Ablation study on context management and training strategies.
        We report the Average Steps per trajectory (as a proxy for exploration depth) and the Success Rate on SWE-bench-Verified. w/o CR denotes the ablation setting where the \textit{Compression Reward} is excluded during RLVR.
    }
    \vspace{-10pt}  
    \label{tab:ablation_metrics}
\end{table}

\vspace{5pt}
\noindent\textbf{Lost-in-the-Middle.}
Comparing Rows (2) and (3) confirms the ``Lost-in-the-Middle'' phenomenon discussed in the Sec~\ref{sec:intro}. Retaining full reasoning history (Interleaved Thinking) significantly degrades performance (12.42\%) compared to discarding it (Current-Step Thinking)(15.83\%), despite the richer context. Additionally, we hypothesize that this degradation may also be partially related to the post-training process of Qwen3 models.
On the contrary, while the average number of steps is more compared to Current-Step Thinking, \ours paradigm ensures that the context length remains manageable. This allows the 8B model to handle long-horizon tasks without suffering from the ``Lost-in-the-Middle'' phenomenon.

\vspace{5pt}
\noindent\textbf{Shallow Reasoning SFT Data Degrades Reasoning Capability.}
Paradoxically, SFT on original 2.2k SWE-Dev trajectories (Row 4) performs worse (14.83\%) than the base model (15.83\%). 
This indicates that training on data restricted to shallow reasoning traces actually constrains the model: it learns to align with the superficial heuristics of the dataset rather than leveraging its full pre-trained potential for deep problem-solving.

We decompose the model response into Reasoning, Textual Summary and Action. For our method, Textual Summary corresponds to the Reasoning Digest. For the Current-Step baseline, represents the standard Thought trace used to justify the action.

\begin{figure}[t]
    \centering
    \includegraphics[width=0.95\linewidth]{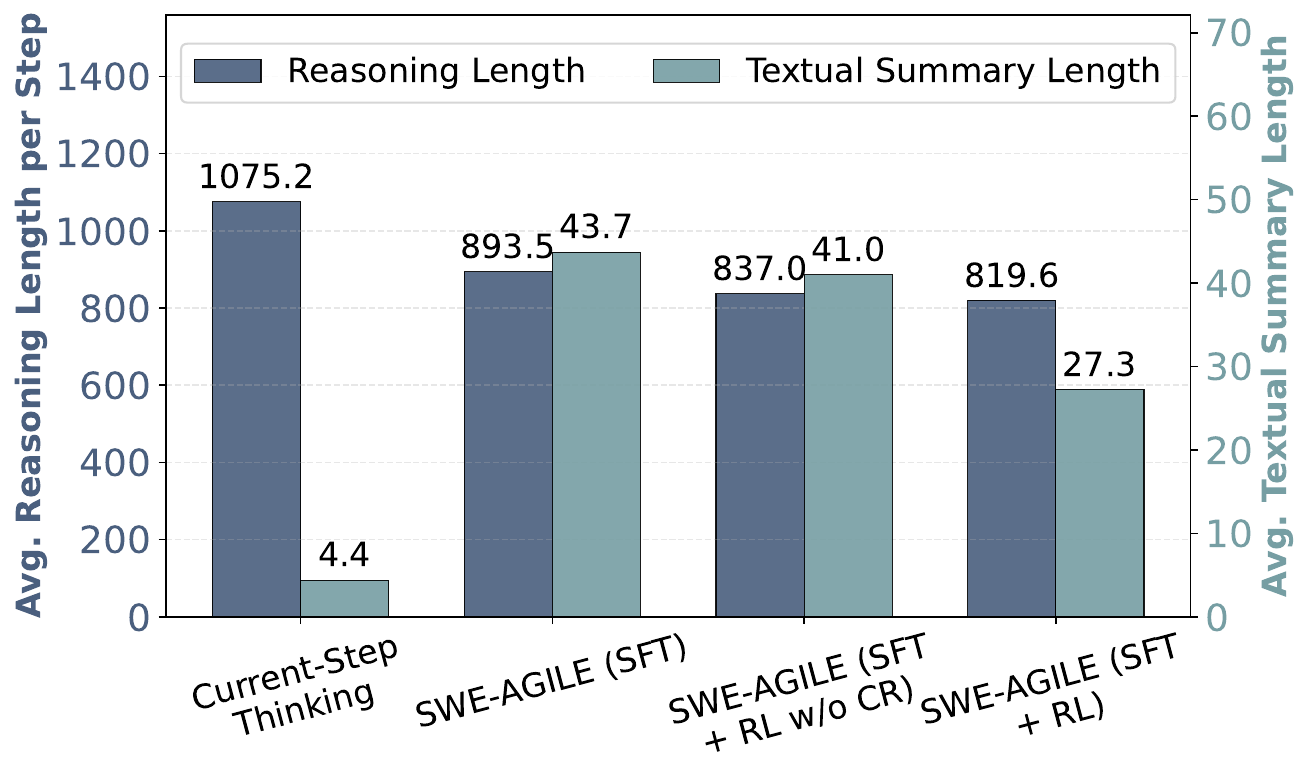}
    \vspace{-10pt}  
    \caption{Average Response Length per Step.}
    \vspace{-10pt}  
    \label{fig:response_decomposition}
\end{figure}

\vspace{5pt}
\noindent\noindent\textbf{Efficiency of \ours.}
To understand how \ours achieves efficiency, we decompose models' average per-step response into Active Reasoning ($r_t$) and Textual Summary ($d_t$) in Fig.~\ref{fig:response_decomposition}.

A key observation is that Current-Step Thinking generates the most verbose active reasoning ($\sim$1,075 tokens/step). Lacking access to enough information of the previous thought, the agent suffers from contextual amnesia and is forced to redundantly reconstruct the state from raw observations at every turn. 
In contrast, \ours (SFT+RL) significantly reduces this load to $\sim$819.6 tokens/step ($-28\%$). This confirms that our Reasoning Digests and Sliding Reasoning Window act as an effective cognitive cache, allowing the agent to perform \emph{incremental reasoning} rather than redundant re-analysis.

While our current paradigm implicitly reduces redundant state reconstruction, a highly promising direction to strictly enforce this efficiency is to quantitatively monitor the reasoning content. By \emph{calculating the embedding similarity between consecutive reasoning steps or employing an LLM-as-a-Judge}, future iterations can explicitly filter out repetitive SFT trajectories or design targeted RLVR penalties, pushing the boundary of cognitive efficiency even further.

\vspace{5pt}
\noindent\textbf{Effect of Compression Reward.}
Comparing \textit{RL w/o Compression Reward} and \textit{RL}, we can see that the Compression Reward reduces the Textual Summary ($d_t$) by $33.4\%$ ($41 \to 27.3$ tokens), while maintaining a comparable task success. As tool execution outputs also consume a large portion of the context window in agentic tasks, tool execution is essential to SWE tasks and the CR mechanism can be readily extended to compress verbose tool outputs. Although saving 13.7 tokens per step seems small in isolation, it represents a 33.4\% relative reduction in the reasoning digest. When combined with tool output compression, the agent could have significantly more interaction turns before hitting context limits, thereby compressing more tokens in reasoning digests.

\vspace{5pt}
Overall, the above experimental results and findings demonstrate that \ours successfully decouples reasoning depth from context cost, achieving the Pareto frontier of performance and efficiency.

\section{Related Work}
Recent advancements in LLMs have spurred the rapid development of autonomous agents across diverse domains, including GUI agents~\citep{lian2025uiagileadvancingguiagents, wang2026offpolicyonpolicyenhancinggui, zhang2026infinitewebscalablewebenvironment, wang2025fedmabenchbenchmarkingmobileagents, gan2026androidcoachimproveonline, chen2025guishepherdreliableprocessreward}, search agents~\citep{tang2025turnlimitstrainingdeep, miromindteam2026mirothinker17h1heavyduty}, optimization techniques for agent tasks acceleration~\citep{huang2026speceyesacceleratingagenticmultimodal}, alongside various other emerging agentic frameworks~\citep{yang2026tooltree, wang2025mcpflowfacilitatingllmagents, hu2026contextagentdynamicdiscoursetrees}. In parallel with these advancements, researchers are increasingly focusing on the highly challenging, multi-turn domain of SWE Agent.

\subsection{SWE Agent}

\vspace{5pt}
\noindent\textbf{Agentic SWE Scaffolds and Pipelines.} 
Current approaches for automated software engineering broadly fall into two categories: agentic scaffolds and pipeline-based frameworks.
Agentic scaffolds, such as SWE-agent~\citep{DBLP:conf/nips/YangJWLYNP24} and OpenHands~\citep{DBLP:conf/iclr/0001LSXTZPSLSTL25}, empower LLMs to autonomously solve tasks by actively navigating repositories, editing files, and executing shell commands. To further enhance decision-making, SWE-Search~\citep{DBLP:conf/iclr/AntoniadesOZXGW25} integrates Monte Carlo Tree Search (MCTS) into the agentic loop.
Conversely, pipeline-based frameworks do use agentic paradigm, but applying structured, multi-stage workflows~\citep{DBLP:journals/pacmse/XiaDDZ25, DBLP:journals/corr/abs-2509-23045, tao2025code, wei2025swerl}. Beyond repository-level bug fixing, agentic frameworks are also effectively deployed for automated adversarial testing to expose code vulnerabilities~\citep{DBLP:journals/corr/abs-2602-20213}.

\vspace{5pt}
\noindent\textbf{Environment Curation.}
Addressing the scarcity of high-quality training trajectories and evaluation tasks, recent research focuses on proposing novel pipelines to create SWE environments~\citep{pan2025training, jain2025regym, DBLP:conf/acl/WangHWTD25, yang2025swesmith, DBLP:journals/corr/abs-2506-19290, guo2026swefactoryautomatedfactoryissue}. 

\vspace{5pt}
\noindent\textbf{Training and Inference-Time Scaling}
Building on these data foundations, works such as SWE-Dev, Kimi-Dev~\citep{DBLP:journals/corr/abs-2509-23045}, ENTROPO~\citep{DBLP:journals/corr/abs-2509-12434} and DeepSWE~\citep{deepswe2025} investigate post-training paradigms like Rejection Sampling Fine-tuning, DPO~\citep{DBLP:conf/nips/RafailovSMMEF23}, and GRPO~\citep{DBLP:journals/corr/abs-2402-03300} to effectively enhance the capabilities of SWE Agents. Complementary to training, inference-time scaling strategies have proven effective for boosting performance during deployment.
Approaches utilized in R2E-Gym, SWE-Dev, DeepSWE, Skywork-SWE and ENTROPO leverage inference-time scaling techniques such as verifier-guided Best-of-N selection to maximize the success rate of tasks.

\subsection{CoT Compression}

Recent advancements in LLMs, such as OpenAI o1 and DeepSeek-R1, have improved performance in System-2 reasoning domains like mathematics and programming by harnessing supervised fine-tuning and reinforcement learning techniques to enhance the Chain-of-Thought (CoT) reasoning. While appropriate CoT sequences improve performance, overlong CoT sequences may introduce significant computational overhead and attention dilution, and even trigger cognitive side-effects such as ``narrative overfitting'', where agents synthesize spurious causal relationships to force coherence~\citep{DBLP:journals/tmlr/SuiCWZZYLWZZCH25, DBLP:journals/corr/abs-2601-01685}.
\citep{arora2025training} trains models to produce minimal yet correct CoT by rewarding correctness while penalizing reasoning length to encourage efficient reasoning. 
TokenSkip~\citep{xia-etal-2025-tokenskip} enables LLMs to skip redundant tokens within CoTs with controllable compression ratio.
O1-Pruner~\citep{DBLP:journals/corr/abs-2501-12570} introduces the Length-Harmonizing Reward, combined with a PPO-style loss, to optimize reasoning LLMs by effectively shortening the CoT length.
Without relying on a reference model, DAST~\citep{shen-etal-2025-dast} employs SimPO to fine-tune reasoning LLMs using a constructed length preference dataset.

Recent CoT compression techniques like LightThinker~\citep{zhang-etal-2025-lightthinker} and InftyThink~\citep{DBLP:journals/corr/abs-2503-06692} focus on compressing the internal thinking process within a single turn. 
Differently, \ours addresses the challenge of maintaining cognitive continuity across multi-turn environmental interactions.
Hence, \ours is orthogonal to those CoT compression techniques and can be deployed in conjunction with them.

\section{Conclusion}

In this paper, we introduce \ours, a framework designed to reconcile reasoning depth with context constraints in the long-horizon SWE task. By integrating a Dynamic Reasoning Context supported by trajectory snapshot training and compression-aware RLVR, \ours enables agents to leverage explicit System-2 reasoning while preventing context explosion.

Crucially, by effectively decoupling transient reasoning overhead from permanent context retention, this framework lays a solid groundwork for explicitly identifying and minimizing redundant re-analyzing as discussed in Efficiency of SWE-AGILE in Sec.~\ref{sec:experiment} , thereby opening new avenues for optimizing agent cognitive efficiency in future research.
\section*{Limitation}

In current version of \ours, the size of the sliding reasoning window is set to a random integer between $[2, 5]$ during backfilling, SFT, RLVR and inference. 
Although this setting reveals the robustness of \ours, more analysis of the size of the sliding reasoning window remains unexplored.


\bibliography{ref}

\appendix

\section{Scaffold}
\label{app:scaffold}

Following R2E-Gym, we use four tools to enable the agent to perform diverse SWE tasks; 1) file editor: for viewing and editing files, 2) search tool: for searching a relevant term in a given file or folder, 3) execute bash: allowing execution of non-interactive bash commands (e.g., for running test scripts), 4) submit: for ending the current trajectory while returning expected outputs.

SWE-Dev trajectories use only three tools, which are basically the same R2E-Gym but lack 2) search tool, although using the bash tool can basically achieve the same effect of a search tool. Therefore, we further collect about 200 trajectories on tasks from R2E-Gym using the four tools to supplement SFT data.

\begin{figure*}[t]
    \centering
    \includegraphics[width=0.95\linewidth]{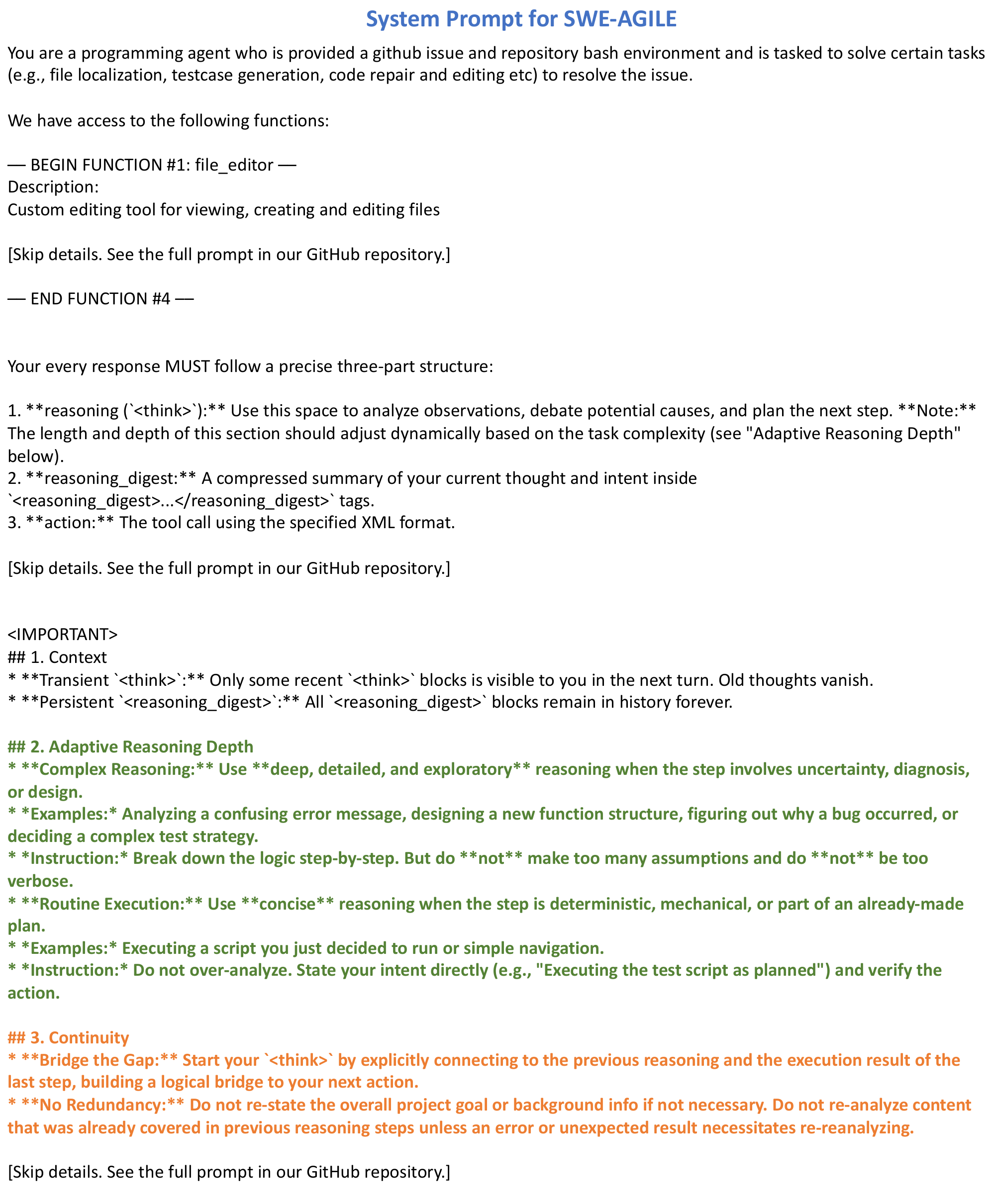}
    \caption{The System Prompt for SWE-AGILE. Coherent with Backfilling Prompt, it helps reducing redundant re-analyzing.}
    \label{fig:agent_prompt}
\end{figure*}

\begin{figure*}[t]
    \centering
    \includegraphics[width=0.95\linewidth]{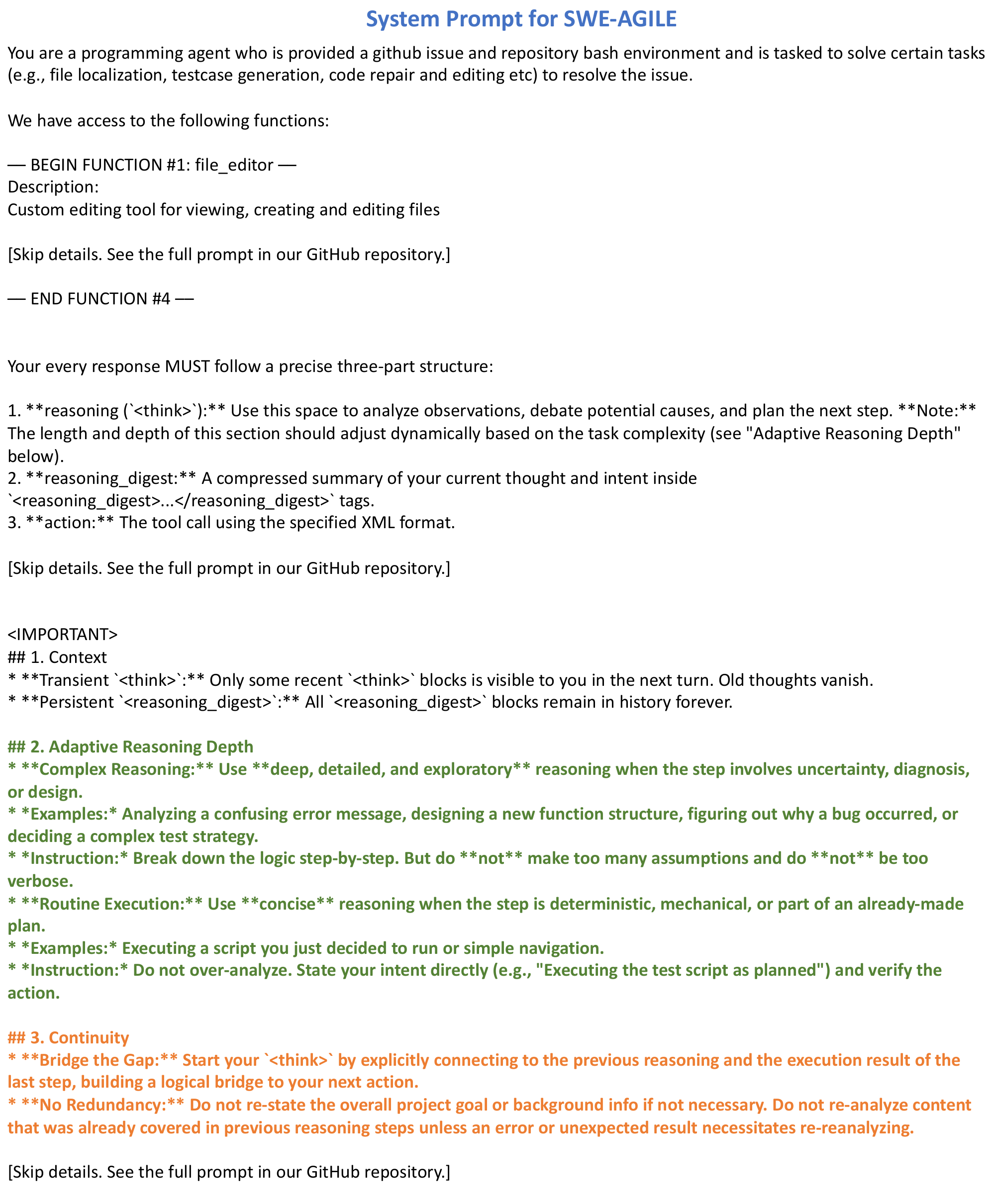}
    \caption{The Prompt for Backfilling Reasoning and Digest. Coherent with Agent System Prompt, it helps reducing redundant re-analyzing.}
    \label{fig:backfilling_prompt}
\end{figure*}

The detailed prompts are provided in Fig.~\ref{fig:agent_prompt} and Fig.~\ref{fig:backfilling_prompt}.

\section{Detailed Hyperparameters}
\label{app:hyper_parameters}

Tabs.~\ref{tab:hyperparams} and~\ref{tab:hyperparams2} provide more detailed hyperparameters. 

\begin{table}[t]
    \centering
    \scalebox{0.9}{
        \begin{tabular}{lc}
            \toprule
            \textbf{Hyperparameter} & \textbf{Value} \\
            \midrule
            Learning Rate & $1 \times 10^{-6}$ \\
            Global Batch Size & 16 \\
            Mini-Batch Size & 8 \\
            Generations per Prompt ($G$) & 8 \\
            Max Prompt Length & 28,582 \\
            Max Response Length & 4096 \\
            Clip Ratio (Low / High) & $0.2$ / $0.28$ \\
            KL Coefficient & 0.0 \\
            Compression Reward Weight $\beta$ & 0.2 \\
            Compression Clipping threshold $\gamma$ & 0.55 \\
            Temperature & 1.0 \\
            Repetition Penalty & 1.15 \\
            Total Epochs & 1 \\
            \bottomrule
        \end{tabular}
    }
    \caption{Hyperparameters for RLVR (DAPO) Training}
    \label{tab:hyperparams}
\end{table}

\begin{table}[t]
    \centering
    \scalebox{0.9}{
        \begin{tabular}{lc}
            \toprule
            \textbf{Hyperparameter} & \textbf{Value} \\
            \midrule
            Learning Rate & $1 \times 10^{-5}$ \\
            Batch Size & 32 \\
            Max Sequence Length & 26000 tokens \\
            Total Epochs & 4 \\
            Sliding Window Size & random in [2, 5] \\
            \bottomrule
        \end{tabular}
    }
    \caption{Hyperparameters for Snapshots Training}
    \label{tab:hyperparams2}
\end{table}

\section{Tool Calling Format}
\label{app:toolcall}

Standard JSON-based tool calling poses significant robustness challenges in the SWE  task. Since tool arguments often include code snippets containing strings and special characters, encapsulating this content within a JSON structure requires complex, multi-level escaping. This complexity frequently leads to syntax errors, particularly when model capabilities are limited. To mitigate these nesting and escaping issues, we adopt an XML-based tool calling format, which allows for more robust parsing of raw code content.

\end{document}